\documentclass{article}
\usepackage{spconf}
\usepackage{graphicx}
\usepackage{booktabs}
\usepackage{amsmath}
\usepackage{tabularx}
\usepackage{color}
\usepackage{url}
\usepackage{multirow}
\usepackage{mathrsfs}

\newcommand{\jy}[1]{{\color{blue}{#1}}}
\title{Topic-to-essay generation with knowledge-based content selection}
\name{Jieyong Wang, Chunyao Song*, Yihao Wu}
\address{College of Computer Science, Nankai University, China}
\begin{document}
%\ninept
%
\maketitle
\begin{abstract}
The topic-to-essay generation task is a challenging natural language generation task that aims to generate paragraph-level text with high semantic coherence based on a given set of topic words. Previous work has focused on the introduction of external knowledge, ignoring the insufficient generated text diversity. In order to improve the generation diversity, we propose a novel copy mechanism model with a content selection module that integrates rich semantic knowledge from the language model into the decoder. Furthermore, we introduce the improved prefix tuning method to train the model, enabling it to adapt to varying input complexities. In addition, we have contributed a new Chinese dataset for TEG tasks. Experimental results demonstrate that the proposed model can improve the generated text diversity by 35\% to 59\% compared to the state-of-the-art method, while maintaining a high level of topic consistency.
\end{abstract}


\section{Introduction}
\label{sec:intro}

\begin{figure}
    \centering
    \includegraphics[width=1\linewidth]{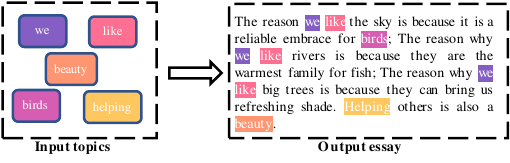}
    \caption{A example for essay generation with given topics.}
    \label{fig:fig_teg}
\end{figure}

Topic-to-essay generation (TEG), which aims at generating fluent, novel, and topic-consistent paragraph-level text with several given topics (keywords), as shown in Fig. \ref{fig:fig_teg}, has a great deal of practical applications. It can be used for automatic advertisement generation, mail generation or keyword-based news writing \cite{leppanen2017data}.

Due to the wide range of applications of TEG, this task has attracted a large amount of research attention. Previous study mentioned that unlike machine translation and text summarization, in TEG task, the semantic resources contained in the input sequence are much smaller than those in the output sequence, and it is almost impossible to generate satisfied text without enough semantic resources \cite{yang}. Therefore, in order to improve the quality of generated texts, researchers have started to explore the introduction of different types of external resources. Some previous studies \cite{yang, Lin, sememe} incorporate knowledge from the commonsense knowledge base like ConceptNet \cite{ConceptNet} and HowNet \cite{hownet}. They utilize different graph structure encoders to fully leverage the external information. However, some studies argue that the information in the training corpus is sufficient to generate high-quality text, so they focus on how to effectively leverage the information within the corpus \cite{PMI-IR, backgroundNetwork}. %And 
Additionally, some other works focus on different aspects, such as the inconsistency between the training and inference processing \cite{relatedEntropy}, the human writing conventions \cite{hierarchical}, and the multiple perspectives of input information \cite{hierarchicalBERT}.
Although these methods produced high-quality text, \textbf{there are still two issues that need to be addressed in TEG. First, the existing methods fail to utilize the rich semantic knowledge in language models, resulting in unsatisfactory text quality. Second, previous methods excessively emphasize 
the importance of labeled text, thereby restricting the diversity of generated text.}

\begin{figure*}[htbp]
    \centering
    \includegraphics[width=1\linewidth]{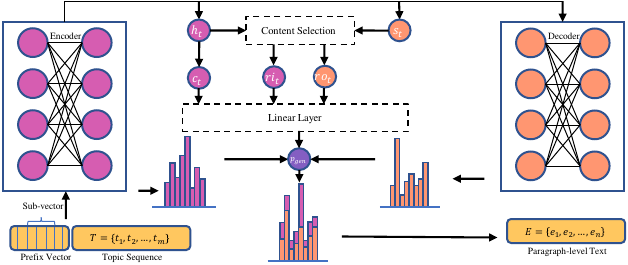}
    \caption{Overview of our proposed model for topic-to-essay generation task.} 
    \label{model.png}
\end{figure*}

To address these problems, we propose a novel content selection topic-to-essay generator based on the encoder-decoder framework. The generator uses a pre-trained GENIUS language model \cite{genius} and a copy mechanism with content selection module, applying a unique training method to protect the knowledge in GENIUS from being destroyed.
GENIUS is trained using a unique sketch reconstruction pre-training task, which enables it to learn knowledge similar to the TEG task while other language models can't. 
To enhance the diversity of generated texts and make full use of the rich semantic knowledge contained in GENIUS, we propose a content selection module to the generator, using the improved prefix tuning method to train. It also monitors the state of the decoder at all time steps to make sure that the generated text is always correlated with the topics. Like prefix tuning \cite{prefix-tuning}, the knowledge in the model is preserved. Furthermore, we have released a new Chinese dataset NAES for the TEG task. This dataset includes news articles and essays on various topics, such as finance, society, and lifestyle. Experimental results demonstrate that our method has an average improvement of over 40\% compared to SOTA methods for text diversity on all evaluation datasets, while maintaining a high level of topic-consistency.

\section{Methodology}
\label{sec:format}
In Figure \ref{model.png}, we present our proposed model, GCS-IPT, which utilizes a pre-trained \textbf{G}ENIUS model comprising of a copy mechanism with \textbf{C}ontent \textbf{S}election module, and is trained using the \textbf{I}mproved \textbf{P}refix-\textbf{T}uning approach.

\subsection{Task Formulation}
The TEG task takes a topic set $T=\{t_1, t_2, ..., t_m\}$ consisting of $m$ topic words as input, aims to generate paragraph-level text $E=\{e_1, e_2, ..., e_n\}$ that is coherent with the input topics and $n$ is the token number of generated text. For example, as shown in Figure \ref{fig:fig_teg}, when given five topics: ``we'', ``like", ``birds", ``helping", and ``beauty", a TEG model will generate paragraph-level text as shown in the figure.

\subsection{Copy mechanism with Content Selection
}
The copy mechanism allows for additional attention to be given to the input sequence during the generation process:
\begin{equation}
    c_t = \sum_{i} \alpha_i^th_i 
\end{equation}
\begin{equation}
p_{gen} = \sigma(W[c_t, s_t] + b)
\end{equation}
At time step $t$, the context vector $c_t$ is calculated with the hidden state of the encoder $h_t$ and the attention of the target sequence to source sequence $\alpha^t$. And we get $p_{gen}$, the probability of selecting a word from the vocabulary where $W$ and $b$ are learnable hyperparameters in a fully-connected layer.

While the copy mechanism enables heightened attention to the topic words during generation, it fails to consider the accurate expression of diverse topic semantics in the resulting text. In order to address this, we propose the Content Selection module to compute the probability $p_{gen}$.

We use the cosine similarity $sim()$ to calculate %calculating 
the similarity $r_t$ between the hidden states of the encoder and the decoder. Then an auxiliary function $f(r_t, v)$ is set to filter information from vector $v$ based on the similarity $r_t$, retaining only the information that surpasses a certain threshold $\phi$.
\begin{equation}
r_t = sim(h_t, s_t)
\end{equation}
\begin{equation}
f(r_t, v)=\begin{cases}
    v, & r_i > \phi, \\
    0, & else .
\end{cases}
\end{equation}

Next, we select the helpful information for generation.
\begin{equation}
    ri_{t} = f(r_t, o(h_t)) × h_t
\end{equation}
\begin{equation}
    ro_t = f(1 - r_t, o(s_t)) × s_t
\end{equation}
The function $o()$ represents a normalization function that scales the vector's maximum value to 1 and the minimum value to 0. The vector $ri_t$ represents the result of content selection on the encoder state, where its non-zero elements indicate the most relevant information from the input topic in the encoder. The vector $ro_t$ represents the result of content selection on the decoder state, selecting information based on the complement of the similarity $r_t$, ensuring that the decoding process takes into account the semantic aspects of the topic that have not been expressed yet.

Finally, the probability $p_{gen}$ is calculated as follows:

\begin{equation}
    p_{gen} = \sigma(W[c_t, ri_t, ro_t] + b)
\end{equation}

Similar to the traditional copy mechanism, the probability distribution for copying words from the input source sequence $P_{copy}$ can be obtained through attention $\alpha$. For the current word $e_t$ to be generated, the words in the source sequence with higher attention scores for $e_t$ are more likely to be copied. Finally, the probability distribution for generating a word $e$ combines the probability distribution from the vocabulary and the probability distribution for copying words.
\begin{equation}
    P(e) = p_{gen}P_{vocab}(e) + (1-p_{gen})P_{copy}(e)
\end{equation}

\subsection{Improved Prefix-tuning}

%\jy{During the training process, we do not want to compromise the semantic knowledge in GENIUS, so we choose to train it using the prefix-tuning method. Considering the complexity of the input text varies in a regular pattern in TEG task, increasing linearly with the number of topic words, we propose the improved prefix-tuning method.}
Throughout the training process, preserving the semantic knowledge within GENIUS is of utmost importance. Therefore, we opt for training it utilizing the prefix-tuning approach. Considering the complexity of the input text often varies in TEG task, which escalating proportionally with the count of topic words, we propose an enhanced variant of the prefix-tuning technique.
%In TEG task, the complexity of the input text varies in a regular pattern, increasing linearly with the number of topic words. To better match this phenomenon, we propose the imporved prefix-tuning method. 

Specifically, prefix vector is a fixed-length vector, but different subvectors are trained based on the number of topics. Let $n$ represents the given number of topics. We obtain the initial index $idx$ for the subvector by applying a linear transformation to the initial state $h_0$ of the encoder. And a softmax normalization is applied to obtain probabilities. Finally, once the initial index and length of the subvector are determined, the selection of the subvector is also finalized.
% \begin{equation}
%     b_{l} = base * n
% \end{equation}
% \begin{equation}
%     idx = argmax(Softmax(\sigma(Wh_0 + b))
% \end{equation}
% \begin{equation}
%     subvector = v[idx: idx + b_l]
% \end{equation}
%%% 

\subsection{Training}

Given a set of topic words as input, a variable-length prefix is added to the front and the model then calculates the state and attention. At each time step of the generation stage, the states of both encoder and decoder are passed to the content selection module. This module computes the representation of the input topics and eventually generates a copying probability distribution. Finally, copying distribution integrates with the vocabulary probability distribution, generating the probability distribution for the output at the current time step. The objective function of the entire training process is to minimize the negative log-likelihood loss.

% \begin{equation}
%     \mathcal{L}_{\mathrm{NLL}} = -\sum_{i=1}^n logp_\theta (e_i|T, e_{<i})
% \end{equation}

\section{Experiments}

\begin{table*}[t]
\resizebox{\textwidth}{!}{
\begin{tabular}{ccccccc}
\hline
Datasets               & Metric      & MTA   & CTEG  & SCTKG(SOTA)          & GENIUS & GCS-IPT           \\ \hline
\multirow{4}{*}{ZHIHU} & BLEU        & 7.09 (+6.63\%)  & 9.72 (-22.22\%)  & \textbf{11.02 (-31.40\%)} & 5.96 (+26.85\%)   & 7.56           \\
                       & DIST-2      & 11.70 (+167.35\%) & 20.49 (+52.66\%) & 23.07 (+35.59\%)          & 27.33 (+14.45\%)  & \textbf{31.28} \\
                       & Consistency & 25.73 (+72.02\%) & 39.42 (+12.28\%) & 43.84 (+0.96\%)          & 40.08 (+10.43\%)  & \textbf{44.26} \\
                       & Novelty     & 70.68 (+20.03\%) & 75.71 (+12.06\%) & 79.54 (+6.66\%)          & 81.83 (+3.68\%)  & \textbf{84.84} \\ \hline
\multirow{4}{*}{ESSAY} & BLEU        & 5.33 (+27.02\%)  & 7.36 (-8.02\%)  & \textbf{9.83 (-31.13\%)}  & 5.17 (+30.95\%)   & 6.77           \\
                       & DIST-2      & 8.64 (+240.74\%)  & 19.33 (+52.30\%) & 19.91 (+47.87\%)          & 27.36 (+7.60\%)  & \textbf{29.44} \\
                       & Consistency & 19.36 (+100.21\%) & 35.68 (+8.63\%) & 37.20 (+4.19\%)          & 36.52 (+6.13\%)  & \textbf{38.76} \\
                       & Novelty     & 67.81 (+21.80\%) & 71.02 (+16.29\%) & 73.65 (+12.14\%)          & 80.16 (+3.03\%)  & \textbf{82.59} \\ \hline
\multirow{4}{*}{NAES}  & BLEU        & 4.81 (+49.27\%)  & 5.26 (+36.50\%)  & \textbf{7.32 (-1.91\%)}  & 7.02 (+2.28\%)   & 7.18           \\
                       & DIST-2      & 8.53 (+247.13\%)  & 14.28 (+107.35\%) & 18.60 (+59.19\%)          & 25.69 (+15.26\%)  & \textbf{29.61} \\
                       & Consistency & 17.62 (+122.42\%) & 26.38 (+48.56\%) & 32.05 (+22.28\%)          & 33.62 (+16.57\%)  & \textbf{39.19} \\
                       & Novelty     & 67.22 (+18.30\%) & 71.20 (+11.69\%) & 72.39 (+9.85\%)          & 74.80 (+6.31\%)  & \textbf{79.52} \\ \hline
\end{tabular}
}
\caption{\label{tab:all}
Experimental results. All evaluation metrics, except for BLEU, outperform baseline models. The numbers in parentheses represent the percentage increase in performance that our method achieves compared to the baseline method.}
\end{table*}

\subsection{Datasets}

\noindent{\textbf{Public Available Dataset: }} We conduct experiments on the ZHIHU and ESSAY \cite{feng} public datasets. The number of topics used follows the same settings as \cite{yang} and the training set and test set are set to 25,000 and 2,000, respectively.

\noindent{\textbf{Self-Constructed Dataset: }}
 Due to the lack of diversity caused by the uniform style of open-source datasets for TEG, we contribute a new Chinese dataset News and Essays (NAES)\footnote{\url{https://mega.nz/folder/5r12GIBb#UF1v6p50CwJOpI4tRssMxQ}} for the TEG task. This dataset includes topic news such as finance, social, culture and tourism, as well as high-scoring essays. We gather articles from publicly accessible online sources, choosing paragraphs that fall within the length range of 50 to 200 words. For extracting key phrases, we employ the YAKE algorithm \cite{YAKE} to extract 5 topics from each text. In addition, we select the top 100 most frequent topics from NAES. The training and test set is set to 18000 and 1400 data samples.

\subsection{Evaluation Setup}

\textbf{Baselines} \\
 We select MTA \cite{feng}, CTEG \cite{yang}, SCTKG \cite{Lin} and GENIUS \cite{genius} as baselines. The first three models are specifically designed for TEG tasks and the SCTKG is the SOTA model. We draw inspiration from \cite{mderank} and \cite{ENN}, where the pre-training tasks of language models have a significant impact on the performance of downstream tasks. The knowledge in GENIUS learned from the sketch reconstruction task \cite{genius} is similar to what TEG tasks require. So we choose GENIUS as one of the baselines for TEG tasks. All experimental results are the average values of five experiments.

\noindent{\textbf{Implement Details}}\\
We utilize genius-base-chinese\footnote{\url{https://huggingface.co/beyond/genius-base-chinese}.} as the initialization parameters for our model. We implement changes to the model structure using PyTorch. Training is conducted for 50 epochs on a single 1080 Ti GPU. The Adam optimizer is employed, with an initial learning rate of 5e-6. During testing, a beam search decoding strategy is applied with a beam width set to 3. For model hyperparameters, we set prefix base length to 30, and the similarity threshold is set to 0.2.

\noindent{\textbf{Automatic Evaluation}} \\
Similarly, based on previous work, we choose the following evaluation method: BLEU score \cite{bleu}, Diversity (DIST-2) \cite{dist}, Consistency \cite{yang, Lin} sand Novelty \cite{yang, Lin}. 
BLEU can reflect the similarity between generated text and labels, so \jy{a} higher BLEU score may limit diversity. However, considering that BLEU can to some extent reflect the quality of text, we still report this metric as a reference. Dist-2 and Novelty both reflect the diversity of generated text. Consistency, on the other hand, indicates whether the generated content is centered around the semantic of the input topic.

\subsection{Performance Comparsion}

The experimental results of automatic evaluation are presented in Table \ref{tab:all}. The results demonstrate that our method achieves the best performance on all metrics except the BLEU score in both the ZHIHU and ESSAY datasets. Particularly, TEG is an open-ended text generation task where the generated content should not heavily rely on the labeled text. At this level, a higher BLEU score may hinder the ability to generate diverse text. However, we still report the BLEU score because it can to some extent reflect the coherence of the generated texts by the model.

Our proposed model maintains an objective BLEU score while surpassing the SOTA method in terms of consistency. This indicates that our approach achieves high-quality text generation with impressive topic-consistency. Furthermore, our method outperforms the SOTA by 35.59\% and 47.87\% on the DIST-2 metric for the two public datasets, and surpasses the SOTA by 6.66\% and 12.14\% on the Novelty. These results demonstrate the superiority of our method in terms of text diversity and novelty compared to the SOTA approaches.

Table \ref{tab:all} also presents the experimental results of several baseline methods compared to our model on NAES. Apart from a slight difference in BLEU compared to SCTKG (state-of-the-art), our model significantly outperforms existing methods in other metrics. Specifically, the baselines exhibit noticeable performance degradation on NAES compared to ZHIHU and ESSAY, whereas our model demonstrates a relatively smaller performance decline. This indicates that our method is more suitable for realistic corpus environments, showcasing stronger adaptability and robustness.

\subsection{Ablation Study}

Table \ref{tab:ab1} presents the ablation experiments. The GCS model represents the addition of a copy mechanism with a content selection module to the GENIUS. GCS-PT denotes the training of GCS using the original prefix-tuning method, while the complete model GCS-IPT refers to the training of GCS using our improved prefix-tuning method.

Due to the incorporation of the copying mechanism and content selection, the model considers not only the probability distribution of the vocabulary during generation but also the probability distribution of the input content. As a result, BLEU and Consistency are improved. With the help of prefix-tuning, which preserving the rich knowledge in language model, all evaluation metrics show improvement. Finally, the improved prefix-tuning method adapts to diverse input information by utilizing prefix vectors of different lengths based on the length of the input topic words. This adaptation enhances the performance in terms of DIST-2 and Novelty.

\begin{table}[]
\setlength{\tabcolsep}{1.7mm}{
\begin{tabular}{ccccc}
\hline
        & \textbf{BLEU} & \textbf{DIST-2} & \textbf{Consistency} & \textbf{Novelty} \\ \hline
GENIUS  & 5.96          & 27.33           & 40.08                & 81.83            \\
GCS     & 6.83          & 29.5            & 41.85                & 82.61            \\
GCS-PT  & 7.63          & 30.39           & 43.91                & 83.21            \\
GCS-IPT & 7.56          & 31.28           & 44.26                & 84.84            \\ \hline
GENIUS  & 5.17          & 27.36           & 36.52                & 80.16            \\
GCS     & 5.78          & 27.66           & 37.10                & 80.66            \\
GCS-PT  & 6.03          & 28.38           & 38.73                & 81.73            \\
GCS-IPT & 6.77          & 29.44           & 38.76                & 82.59            \\ \hline
\end{tabular}
}
\caption{\label{tab:ab1}
Ablation study on ZHIHU (up) and ESSAY (down).
}
\end{table}

% \begin{table}[]
% \setlength{\tabcolsep}{0.9mm}{
% \begin{tabular}{ccccc}
% \hline
%         & \textbf{BLEU} & \textbf{DIST-2} & \textbf{Consistency} & \textbf{Novelty} \\ \hline
% GENIUS  & 5.17          & 27.36           & 36.52                & 80.16            \\
% GCS     & 5.78          & 27.66           & 37.10                & 80.66            \\
% GCS-PT  & 6.03          & 28.38           & 38.73                & 81.73            \\
% GCS-IPT & 6.77          & 29.44           & 38.76                & 82.59            \\ \hline
% \end{tabular}
% }
% \caption{\label{tab:ab2}
% Ablation study on ESSAY dataset.
% }
% \end{table}

\section{Conclusion}
In this paper, we present a novel content selection module within a topic-to-essay model that incorporates a copy mechanism which integrates %integrate 
rich semantic knowledge in language model to the generation process. Additionally, we introduce an improved prefix tuning training process to further enhance the model's performance by allowing the model to adapt to both simple and complex topic inputs. In addition, we contribute a large Chinese TEG task dataset that has multi-topic text.  %such as finance, society, culture and so on. 
Experiments show that our model can generate more diverse, novel text while maintaining a high topic-consistency, and notably outperform other baselines in text diversity. 

\vfill\pagebreak

\newpage

\bibliographystyle{IEEEbib}
\bibliography{strings,refs}

\begin{thebibliography}{10}

\bibitem{leppanen2017data}
Leo Lepp{\"a}nen, Myriam Munezero, Mark Granroth-Wilding, and Hannu Toivonen,
\newblock ``Data-driven news generation for automated journalism,''
\newblock in {\em Proceedings of the 10th international conference on natural language generation}, 2017, pp. 188--197.

\bibitem{yang}
Pengcheng Yang, Lei Li, Fuli Luo, Tianyu Liu, and Xu~Sun,
\newblock ``Enhancing topic-to-essay generation with external commonsense knowledge,''
\newblock in {\em Proceedings of the 57th annual meeting of the association for computational linguistics}, 2019, pp. 2002--2012.

\bibitem{Lin}
Lin Qiao, Jianhao Yan, Fandong Meng, Zhendong Yang, and Jie Zhou,
\newblock ``A sentiment-controllable topic-to-essay generator with topic knowledge graph,''
\newblock {\em arXiv preprint arXiv:2010.05511}, 2020.

\bibitem{sememe}
Dan Luo, Xinyi Ning, and Chunhua Wu,
\newblock ``Sememe-based topic-to-essay generation with neural networks,''
\newblock in {\em Journal of Physics: Conference Series}. IOP Publishing, 2021, vol. 1861, p. 012068.

\bibitem{ConceptNet}
Robyn Speer, Joshua Chin, and Catherine Havasi,
\newblock ``Conceptnet 5.5: An open multilingual graph of general knowledge,''
\newblock in {\em Proceedings of the Thirty-First {AAAI} Conference on Artificial Intelligence, February 4-9, 2017, San Francisco, California, {USA}}, Satinder Singh and Shaul Markovitch, Eds. 2017, pp. 4444--4451, {AAAI} Press.

\bibitem{hownet}
Jingwen Cao, Tiexin Wang, Wenxin Li, and Chuanqi Tao,
\newblock ``A method of calculating the semantic similarity between english and chinese concepts,''
\newblock in {\em Machine Learning and Intelligent Communications: 4th International Conference, MLICOM 2019, Nanjing, China, August 24--25, 2019, Proceedings 4}. Springer, 2019, pp. 313--324.

\bibitem{PMI-IR}
Xinyi Ning,
\newblock ``Topic-to-text generation with pmi-ir additional semantic information,''
\newblock in {\em 2021 International Conference on Asian Language Processing (IALP)}. IEEE, 2021, pp. 131--136.

\bibitem{backgroundNetwork}
Dan Luo, Xinyi Ning, Chunhua Wu, Maonan Wang, and Jing Wu,
\newblock ``Topic-to-essay generation with corpus-based background information,''
\newblock in {\em Journal of Physics: Conference Series}. IOP Publishing, 2021, vol. 1827, p. 012127.

\bibitem{relatedEntropy}
Zhiyue Liu, Jiahai Wang, and Zhenghong Li,
\newblock ``Topic-to-essay generation with comprehensive knowledge enhancement,''
\newblock in {\em Machine Learning and Knowledge Discovery in Databases. Applied Data Science Track: European Conference, ECML PKDD 2021, Bilbao, Spain, September 13--17, 2021, Proceedings, Part V 21}. Springer, 2021, pp. 302--318.

\bibitem{hierarchical}
Wangbo He and Yuan Rao,
\newblock ``Transformer-based hierarchical topic-to-essay generation,''
\newblock {\em Procedia Computer Science}, vol. 202, pp. 414--421, 2022.

\bibitem{hierarchicalBERT}
Fuqiang Lin, Xingkong Ma, Yaofeng Chen, Jiajun Zhou, and Bo~Liu,
\newblock ``Pc-san: Pretraining-based contextual self-attention model for topic essay generation,''
\newblock {\em KSII Transactions on Internet and Information Systems (TIIS)}, vol. 14, no. 8, pp. 3168--3186, 2020.

\bibitem{genius}
Biyang Guo, Yeyun Gong, Yelong Shen, Songqiao Han, Hailiang Huang, Nan Duan, and Weizhu Chen,
\newblock ``Genius: Sketch-based language model pre-training via extreme and selective masking for text generation and augmentation,''
\newblock {\em arXiv preprint arXiv:2211.10330}, 2022.

\bibitem{prefix-tuning}
Xiang~Lisa Li and Percy Liang,
\newblock ``Prefix-tuning: Optimizing continuous prompts for generation,''
\newblock in {\em Proceedings of the 59th Annual Meeting of the Association for Computational Linguistics and the 11th International Joint Conference on Natural Language Processing (Volume 1: Long Papers)}, 2021, pp. 4582--4597.

\bibitem{feng}
Xiaocheng Feng, Ming Liu, Jiahao Liu, Bing Qin, Yibo Sun, and Ting Liu,
\newblock ``Topic-to-essay generation with neural networks.,''
\newblock in {\em IJCAI}, 2018, pp. 4078--4084.

\bibitem{YAKE}
Ricardo Campos, V{\'{\i}}tor Mangaravite, Arian Pasquali, Al{\'{\i}}pio Jorge, C{\'{e}}lia Nunes, and Adam Jatowt,
\newblock ``Yake! keyword extraction from single documents using multiple local features,''
\newblock {\em Inf. Sci.}, vol. 509, pp. 257--289, 2020.

\bibitem{mderank}
Linhan Zhang, Qian Chen, Wen Wang, Chong Deng, Shiliang Zhang, Bing Li, Wei Wang, and Xin Cao,
\newblock ``Mderank: A masked document embedding rank approach for unsupervised keyphrase extraction,''
\newblock {\em arXiv preprint arXiv:2110.06651}, 2021.

\bibitem{ENN}
Anton Sinitsin, Vsevolod Plokhotnyuk, Dmitriy Pyrkin, Sergei Popov, and Artem Babenko,
\newblock ``Editable neural networks,''
\newblock {\em arXiv preprint arXiv:2004.00345}, 2020.

\bibitem{bleu}
Kishore Papineni, Salim Roukos, Todd Ward, and Wei-Jing Zhu,
\newblock ``Bleu: a method for automatic evaluation of machine translation,''
\newblock in {\em Proceedings of the 40th annual meeting of the Association for Computational Linguistics}, 2002, pp. 311--318.

\bibitem{dist}
Jiwei Li, Michel Galley, Chris Brockett, Jianfeng Gao, and Bill Dolan,
\newblock ``A diversity-promoting objective function for neural conversation models,''
\newblock {\em arXiv preprint arXiv:1510.03055}, 2015.

\end{thebibliography}

\end{document}